\documentclass[10pt,twocolumn,letterpaper]{article}

\usepackage{cvpr}
\usepackage{times}
\usepackage{epsfig}
\usepackage{graphicx}
\usepackage{amsmath}
\usepackage{amssymb}
\usepackage{booktabs}
\usepackage{multirow}
\usepackage{makecell}
\usepackage{xcolor}
\usepackage{microtype}
\usepackage{enumitem}
\usepackage{subcaption}

\graphicspath{
  {figures/}
  {../badas-1.5-article/figures/}
  {../../results/results_summary/figures/}
}

\definecolor{cvprblue}{rgb}{0.21,0.49,0.74}
\usepackage[pagebackref,breaklinks,colorlinks,allcolors=cvprblue]{hyperref}

\newcommand{\badas}{BADAS-2.0}
\newcommand{\flash}{BADAS-2.0-Flash}
\newcommand{\flashlite}{BADAS-2.0-Flash-Lite}
\newcommand{\reason}{BADAS-Reason}
\newcommand{\badasone}{BADAS-1.0}

\begin{document}

\title{Beyond the Beep: Scalable Collision Anticipation and\\
Real-Time Explainability with BADAS-2.0}

\author{Roni Goldshmidt \quad Hamish Scott \quad Lorenzo Niccolini \quad Hernan Matzner\\
Nexar AI\\
{\tt\small \{roni.goldshmidt, hamish.scott, lorenzo.niccolini, hernan.matzner\}@getnexar.com}
}
\date{}

\begin{abstract}
We present \badas{}, the second generation of our collision
anticipation system, building on \badasone{}~\cite{badas10}, which established
that fine-tuning V-JEPA2~\cite{vjepa2} on large-scale, ego-relevant dashcam
data dramatically outperforms both academic baselines and ADAS
systems. \badas{} advances the state of the art along three coherent axes.

\textbf{(i)~Long-tail benchmark and accuracy:} We introduce a new 10-group
long-tail benchmark targeting rare and safety-critical driving scenarios ---
the primary testbed for evaluating model robustness on underrepresented
collision types. To populate it, we employ an pipeline in
which \badasone{} serves as a deployed active oracle, continuously scoring
millions of unlabeled drives to surface high-risk and informative candidates
for annotation. Combined with Nexar's geospatial Atlas platform~\cite{nexaratlas}
for targeted harvesting of rare long-tail scenarios, this strategy grows
the training corpus from 40k to \textbf{178,500 labeled videos}
($\sim$\textbf{2M clips}), achieving consistent state-of-the-art gains
across \emph{all} benchmark subgroups, with the most pronounced improvements
on the hardest long-tail categories.

\textbf{(ii)~Knowledge distillation to edge:} Domain-specific
self-supervised pre-training on 2.25M unlabeled driving videos enables
effective distillation into \flash{} (86M) and \flashlite{} (22M) ---
delivering a \textbf{7--12$\times$ runtime speedup} while retaining
near-parity accuracy, making real-time collision anticipation practical
for edge deployment.

\textbf{(iii)~Explainability:} \badas{} produces real-time, object-focused
attention heatmaps that localize which scene elements drive each prediction.
\reason{}~\cite{qwen3vl} extends this into language: a fine-tuned
vision-language model that ingests the last video frame together with its
heatmap and generates the required driver action alongside structured
textual reasoning for avoiding the detected hazard.

Inference code and evaluation benchmarks are publicly available.
\end{abstract}

\begin{cvprteaser}
  \centering
  \includegraphics[width=\linewidth]{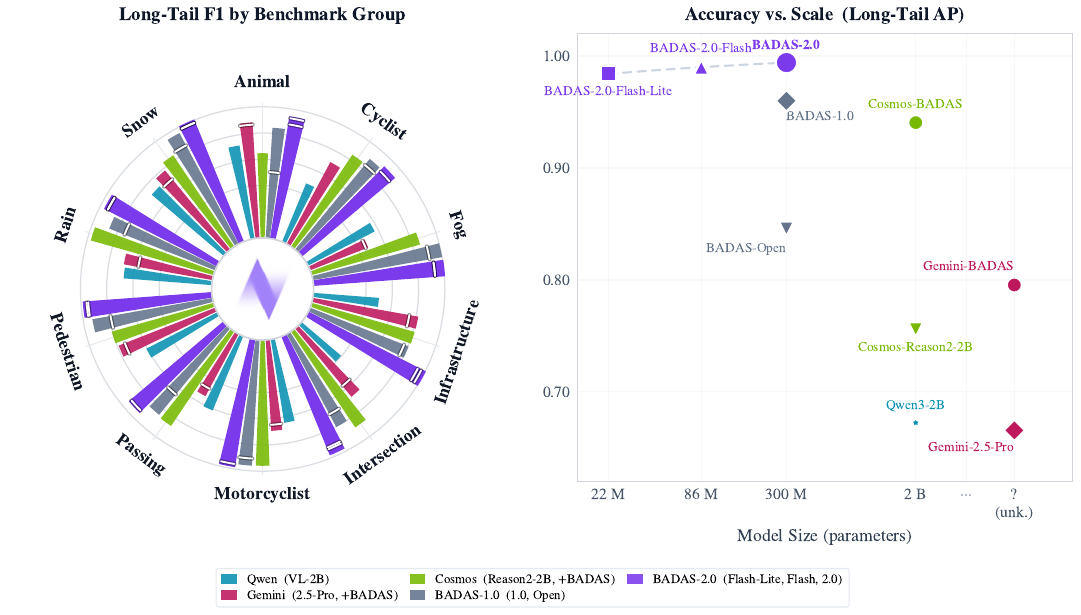}
  \captionof{figure}{\textbf{Long-tail benchmark: per-group F1 (\emph{left}) and AP vs.\ model scale (\emph{right}).}
    \emph{Left}: Radial bars show F1 across 10 scenario groups; each family is color-coded.
    \badas{} (purple) leads every group.
    It can be seen that autoregressive VLM models, even after fine-tuning on the BADAS dataset (the BADAS versions of Cosmos/Gemini), achieve significantly lower performance than the VJEPA2-based BADAS model.
    \emph{Right}: Long-tail AP vs.\ model size.
    The \badas{} family defines an efficient frontier (dashed curve):
    \badas{}-Flash-Lite (22\,M params, AP\,0.984) surpasses Cosmos-BADAS (2\,B params, AP\,0.941)
    at 91$\times$ fewer parameters.}
  \label{fig:radar}
\end{cvprteaser}

\maketitle
\thispagestyle{empty}

\section{Introduction}

Ego-centric collision anticipation — predicting whether the \emph{filming
vehicle} will be involved in a collision — is a distinct and harder task than
general accident detection.  \badasone{}~\cite{badas10} established this
distinction: it re-annotated public benchmarks (DAD~\cite{dad},
DADA-2000~\cite{dada2000}, DoTA~\cite{dota}) to isolate ego-relevant events,
and demonstrated that fine-tuning V-JEPA2~\cite{vjepa2} on 40k ego-centric
dashcam clips outperforms both RNN- and transformer-based accident
detectors~\cite{ustring,dsta} and non-commercial forward-collision warning
systems~\cite{fcwadas}.  The key insight behind this result is that
V-JEPA2's masked spatiotemporal prediction objective learns to represent the
\emph{dynamics} of an unfolding scene — capturing the motion patterns that
precede a collision — whereas prior methods rely on appearance-based features
or semantic priors that transfer poorly to the ego-centric dashcam distribution.

\badas{} addresses three unsolved constraints from \badasone{}: the
\textbf{training corpus} (40k clips) lacked rare long-tail scenarios,
leading to fragile performance; \textbf{inference cost} (2.5\,s/window)
precluded on-device deployment; and the \textbf{scalar risk score} provided
no explanation of which object triggered an alert.

We address all three through an integrated arc: collect \emph{more data,
intelligently}; train a stronger model; distill to edge latency; and add
explainability via heatmaps and VLM fine-tuning.  Each stage builds on the
previous: the dataset is assembled using \badasone{} as an oracle with a
human-in-the-loop (HITL) annotation process; the distilled models start from
SSL weights pre-trained on millions of unlabeled driving videos; and the VLM
pipeline is trained on frames selected by the BADAS attention mechanism.

\paragraph{Contributions.}
\begin{itemize}[nosep,leftmargin=*]

\item \textbf{Intelligent data mining at scale (Sec.~\ref{sec:dataset}).}
We use \badasone{} as an active mining oracle over millions of unlabeled
Nexar drives, supplemented by targeted geospatial queries via Nexar Atlas
for rare scenarios.  The training corpus expands 5$\times$ to 178,500
labeled videos ($\sim$2M windowed training clips), with a new
\textbf{10-group long-tail benchmark} (888 manually verified clips across
animal, cyclist, fog, infrastructure, intersection, motorcyclist,
passing/overtaking, pedestrian, rain, and snow).

\item \textbf{Improved accuracy across all benchmarks
(Sec.~\ref{sec:core}).}  \badas{} (ViT-L, 300M)
improves consistently over \badasone{} on every benchmark, including
Kaggle mAP from 0.925 $\to$ \textbf{0.940}, with gains across all
long-tail scenario groups.

\item \textbf{Knowledge distillation to edge models
(Sec.~\ref{sec:distill}).}  We demonstrate that domain-specific SSL
pre-training on 2.25M unlabeled driving videos transforms a ViT-S from
near-random to near-production quality before any distillation.  Subsequent
KD from the ViT-L teacher yields \flash{} (86M, 4.8\,ms) and \flashlite{}
(22M, 2.8\,ms), both within the real-time budget on all tested platforms.

\item \textbf{BADAS-Reason: Visual and Textual Explainability
(Sec.~\ref{sec:extensibility}).}
BADAS-Reason extends BADAS with a two-pronged explainability layer.
On the visual side, attention maps are extracted at inference time to
highlight the objects and regions that triggered each alert.
On the textual side, we fine-tune Qwen3-VL-4B~\cite{qwen3vl} via
QLoRA~\cite{qlora} on a curated dataset of peak-risk frames paired
with manually annotated descriptions, each specifying the recommended
driver action and the model's reasoning behind the alert.
The resulting model reduces perplexity by 87\% and achieves a
$3.6\times$ improvement in action-match accuracy over the zero-shot
baseline.

\end{itemize}

\section{Related Work}

\paragraph{Collision anticipation.}
Existing methods range from RNN-based adaptive loss functions~\cite{ustring}
to spatial-temporal attention transformers~\cite{dsta} and graph networks
modelling agent relationships~\cite{abductive}.  \badasone{}~\cite{badas10}
showed the primary bottleneck is data alignment — ego-relevant events rather
than general accident footage — rather than architecture.

\paragraph{Active learning and model-assisted mining.}
Self-training pipelines that use model scores to prioritise unlabeled data
for annotation are established in image classification~\cite{pseudolabel}.
We apply this to collision anticipation: \badasone{}'s risk scores surface
high-risk clips for human review, scaling the corpus without exhaustive search.

\paragraph{Knowledge distillation for video.}
Feature- and response-based distillation~\cite{kd} are well-established.
For video, domain-specific SSL pre-training via V-JEPA-style masked
prediction~\cite{vjepa,vjepa2} is a prerequisite: V-JEPA2 was released
only at ViT-L scale, and the domain gap from hand-gesture alternatives is
too large for supervised fine-tuning alone.

\paragraph{VLMs for driving.}
Large VLMs are increasingly applied to driving~\cite{drivelm}.
\reason{} uses a fine-tuned VLM as an explanation generator consuming
BADAS attention outputs, adding natural-language reasoning at real-time speed.

\section{BADAS-2.0 Dataset}
\label{sec:dataset}

\subsection{Intelligent Data Collection}

\badasone{} was trained on $\approx$40k clips.  Its primary weakness was
long-tail coverage: animal crossings, nighttime intersections, and adverse
weather were underrepresented, leading to fragile recall on these groups.
\badas{} assembles 178,500 labeled videos through two strategies executed
after \badasone{} was trained.

\noindent\textbf{Active mining with \badasone{} (Oracle + HITL).}
We run \badasone{} over millions of unlabeled Nexar drives; clips exceeding
a risk threshold are surfaced for human review.  Professional annotators
independently confirm or reject each label — the oracle acts purely as a
retrieval filter, and every clip entering training carries a human-verified
label.  High-confidence false positives are retained as hard negatives to
directly reduce false alarm rates.

\noindent\textbf{Rare-event harvesting via Nexar Atlas~\cite{nexaratlas}.}
Atlas is Nexar's geospatial intelligence platform, which indexes billions of
road miles by scene semantics, weather, road type, time of day, and detected
objects.  Targeted queries extracted clips matching underrepresented
categories — animal crossings, nighttime intersections, rain, snow, fog,
passing/overtaking scenarios — and sent them for expert annotation.  This
directly addresses the long-tail gaps measured in \badasone{}.

\subsection{Annotation and Labeling}

Professional annotators applied a seven-type ontology (collision,
near-collision, hard brake, normal driving, intersection,
passing/overtaking, other), recording direction, third-party involvement,
and event timestamps.

\subsection{Dataset Statistics}

Table~\ref{tab:dataset} summarises the three data generations.

\begin{table}[t]
\centering
\small
\caption{\textbf{BADAS dataset evolution.}  Clips are windowed training
  segments extracted from labeled videos.}
\label{tab:dataset}
\resizebox{\columnwidth}{!}{%
\begin{tabular}{lrrr}
\toprule
 & \textbf{BADAS-Open} & \textbf{BADAS-1.0} & \textbf{BADAS-2.0} \\
\midrule
Labeled videos & 1.5k & 40k & 178.5k \\
Training clips & 15k & 400k & $\sim$2M \\
Positives & 0.8k & 20k & 24.1k \\
Negatives & 0.8k & 20k & 119.5k \\
Hard-braking events & — & — & 34.9k \\
Non-front collisions & — & — & 4.5k \\
Long-tail test groups & — & — & \textbf{10} \\
Long-tail test clips & — & — & \textbf{888} \\
\bottomrule
\end{tabular}}
\end{table}

\subsection{Long-Tail Evaluation Benchmark}

We introduce a \textbf{10-group long-tail benchmark} of 888 manually
verified 9-second clips, covering animal, pedestrian, intersection,
passing/overtaking, cyclist, fog, infrastructure, motorcyclist, rain, and
snow.  Each
clip is a standardized segment placing the annotated event at $t=6$\,s,
leaving 3\,s of post-event footage.  Every clip was independently reviewed
on five quality criteria by human annotators before inclusion.

\section{Improved Core Model}
\label{sec:core}

\subsection{Architecture}

\badas{} fine-tunes a V-JEPA2 ViT-L backbone~\cite{vjepa2} (300M
parameters, $D$=1024, 24 transformer layers) end-to-end on 16-frame clips
at $256{\times}256$ resolution and 8\,fps.  Spatial patch tokens are
mean-pooled across frames to produce a compact clip-level representation.
A future-prediction branch generates the backbone's estimated scene
representation 1\,s ahead and concatenates it with the current-clip
features, giving the prediction head access to both present evidence and
near-future dynamics.  A three-layer GELU MLP (hidden dim 768, dropout 0.1)
maps this combined vector to a collision probability.

Training changes over \badasone{}: (i) gradient checkpointing for
memory-efficient full fine-tuning; (ii) a 2:1 negative-to-positive
oversampling ratio counteracting the $\approx$88\% negative prevalence in the
mined corpus; (iii) cosine learning-rate annealing with label smoothing, gradient
clipping, and mixed-precision training.  The enlarged dataset provides 4.5$\times$
more labeled windows for learning rare scenarios.

\subsection{Main Results}

Table~\ref{tab:main} reports results on the Nexar Kaggle competition
(1,344 clips, single window per video).  Every metric improves over
\badasone{}: Kaggle mAP from 0.925 to 0.940, with the \flash{} variant
achieving the highest mAP (0.941).

\begin{table*}[!t]
\centering
\small
\setlength{\tabcolsep}{5pt}
\caption{\textbf{Main results.}  Nexar Kaggle competition (single window,
  mean AP over three lead-time thresholds).  Best per column in
  \textbf{bold}; second-best \underline{underlined}.
  $\downarrow$\,lower is better for FPR.}
\label{tab:main}
\begin{tabular}{lccc|cc|c}
\toprule
& \multicolumn{3}{c|}{\textbf{Kaggle (AP@TTA)}}
& \multicolumn{2}{c|}{\textbf{Kaggle}}
& \\
\textbf{Model} & @0.5s & @1.0s & @1.5s
& mAP & FPR$\downarrow$
& Params \\
\midrule
\badasone{} (ViT-L)
  & 0.935 & 0.936 & 0.904
  & 0.925 & 10.9\%
  & 300M \\
\midrule
\badas{} (ViT-L)
  & 0.943 & \underline{0.957} & \textbf{0.921}
  & \underline{0.940} & \textbf{4.6\%}
  & 300M \\
\flash{} (ViT-B)
  & \underline{0.945} & \textbf{0.962} & \underline{0.915}
  & \textbf{0.941} & \underline{9.7\%}
  & 86M \\
\flashlite{} (ViT-S)
  & \textbf{0.946} & 0.947 & 0.907
  & 0.933 & 12.2\%
  & 22M \\
\bottomrule
\end{tabular}
\end{table*}

\section{Knowledge Distillation to Edge Models}
\label{sec:distill}

\subsection{The Domain Gap Problem}

Meta released V-JEPA2 only at ViT-L scale; no smaller V-JEPA2 checkpoints
are publicly available.  Training a randomly initialized ViT-S or ViT-B
directly on the BADAS supervision signal fails: without a strong video
prior the backbone never converges to reliable early-warning behavior.
Our SSL ablation (Table~\ref{tab:ssl}) quantifies this gap precisely.

\subsection{Domain-Specific SSL Pre-Training}

Before any supervised training, we pre-train a ViT-B and ViT-S backbone on
\textbf{2.25 million unlabeled Nexar dashcam videos} ($\approx$5M clips)
using V-JEPA-style masked feature prediction~\cite{vjepa2}.  The SSL
objective trains the backbone to predict abstract spatiotemporal representations
of masked video regions from visible context, adapting positional embeddings,
motion representations, and attention patterns to the dashcam distribution.
This phase introduces no labeled data — only the raw video stream.

\subsection{Two-Phase Knowledge Distillation}

After SSL pre-training, we distill from the frozen \badas{} ViT-L teacher into
the domain-adapted student via a composite loss:

\begin{equation}
  \mathcal{L} = \alpha_{\text{hard}}\,\mathcal{L}_{\text{BCE}} +
                \alpha_{\text{logit}}\,\mathcal{L}_{\text{KD}} +
                \alpha_{\text{feat}}\,\mathcal{L}_{\text{feat}}
\end{equation}

\noindent where $\mathcal{L}_{\text{KD}}$ is the KL divergence between
student and teacher soft probability distributions at temperature $\tau=4.0$,
and $\mathcal{L}_{\text{feat}}$ matches intermediate feature representations
(4 selected layers).  Loss weights are $\alpha_{\text{hard}}=0.3$,
$\alpha_{\text{logit}}=0.6$, $\alpha_{\text{feat}}=0.1$.

Training proceeds in two phases over 4,000 steps (effective batch 512, 8 GPUs):
\textbf{Phase 1} (steps 0–3,000) trains with the full composite loss, allowing
the student to absorb the teacher's calibrated uncertainty on borderline cases;
\textbf{Phase 2} (steps 3,000–4,000) drops the teacher and trains only on hard
ground-truth BCE, sharpening probability outputs for deployment.

\subsection{SSL Pre-Training Ablation}

Table~\ref{tab:ssl} isolates the contribution of each stage for ViT-S,
evaluated on the 10-group long-tail benchmark.

\begin{table}[t]
\centering
\small
\caption{\textbf{SSL pre-training ablation (ViT-S).}  Evaluated on the
  10-group long-tail benchmark.  Without SSL, random initialization
  produces near-random AP.  SSL alone approaches production quality; SSL
  plus distillation closes the remaining gap.}
\label{tab:ssl}
\setlength{\tabcolsep}{4pt}
\begin{tabular}{lccccc}
\toprule
\textbf{Model} & \textbf{SSL} & \textbf{KD} & \textbf{AP} & \textbf{F1} & \textbf{FPR$\downarrow$} \\
\midrule
ViT-S (no SSL)   & $\times$ & $\times$ & 0.693 & 0.629 & 35.5\% \\
ViT-S (SSL only) & $\checkmark$ & $\times$ & 0.974 & 0.880 & 20.6\% \\
\flashlite{}     & $\checkmark$ & $\checkmark$ & \textbf{0.984} & \textbf{0.926} & \textbf{9.1\%} \\
\bottomrule
\end{tabular}
\end{table}

\begin{figure*}[t]
  \centering
  \includegraphics[width=\linewidth]{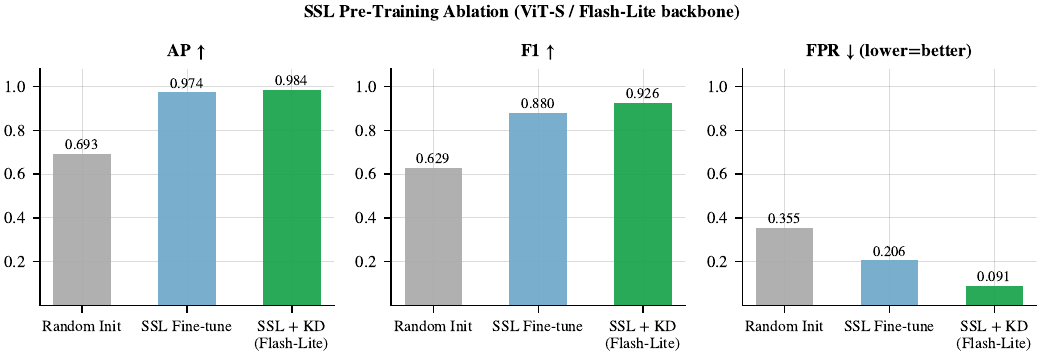}
  \caption{\textbf{SSL pre-training ablation} (ViT-S\,/\,\flashlite{} backbone).
    SSL pre-training alone delivers $+28.1$\,pp AP over random initialisation;
    adding knowledge distillation halves FPR ($20.6\%\to9.1\%$) with a
    further $+1.0$\,pp AP gain.}
  \label{fig:ssl_ablation}
\end{figure*}

Figure~\ref{fig:ssl_ablation} visualises the three-stage progression.
SSL pre-training alone provides \textbf{+28.1\,pp AP} over random
initialization — the dominant effect.  Distillation on top adds
\textbf{+1.0\,pp AP} and, more importantly, \textbf{halves the FPR} from
20.6\% to 9.1\%, as the teacher's calibrated uncertainty transfers to the
student.

\subsection{EWR and MTTA Dynamics Across Model Sizes}

Two complementary metrics capture the temporal behaviour of each model variant.
\textbf{Early Warning Recall (EWR)} measures what fraction of events the model
detects \emph{before} they occur, and follows a clear capacity ordering:
\badas{} (91.3\%) $>$ \flash{} (89.9\%) $>$ \flashlite{} (85.5\%).
Larger models detect a greater share of events, including subtle late-developing
collisions where only brief visual evidence is available; capacity-constrained
models require stronger proximate cues and miss these harder cases entirely.

\textbf{Mean Time to Alert (MTTA)} measures the average lead time
\emph{among detected events only}, and exhibits the opposite ordering:
\flashlite{} (1.46\,s) $>$ \flash{} (1.42\,s) $>$ \badas{} (1.31\,s).
This is a selection effect: smaller models only alert on unambiguous,
visually clear events — which tend to develop over longer time horizons
— while \badas{} also catches harder late-onset events where minimal warning
time is available, lowering its average lead time among detected cases.
The practical implication is that \badas{} provides the most total early
warnings across the broadest range of collision dynamics.

\section{Extensibility: Heatmaps and BADAS-Reason}
\label{sec:extensibility}

\subsection{Training-Free Attention Heatmaps}

BADAS produces spatial attribution heatmaps using a \textbf{training-free}
method that requires no additional modules.  For each prediction window, we
extract self-attention weights from late encoder layers (layers 12–20 for
ViT-L; 8–12 for ViT-B), apply exponential temporal weighting to emphasize
frames close to the alert:
\begin{equation}
  w_t = \frac{\exp\!\bigl((t-7)/T\bigr)}{\sum_{t'}\exp\!\bigl((t'-7)/T\bigr)},
  \quad T = 2.0,
\end{equation}
spatially aggregate to a 16$\times$16 map per layer, average across layers,
upsample to 256$\times$256, and normalize to $[0,1]$.  The only runtime
requirement is eager attention mode (disabling FlashAttention) to expose
weight tensors; a single forward pass suffices.

We evaluate localization via \textbf{Pointing Game Accuracy (PGA)}: whether
the peak activation pixel falls inside a ground-truth bounding box on 1,894
annotated danger clips.

\begin{figure}[t]
  \centering
  \includegraphics[width=\linewidth]{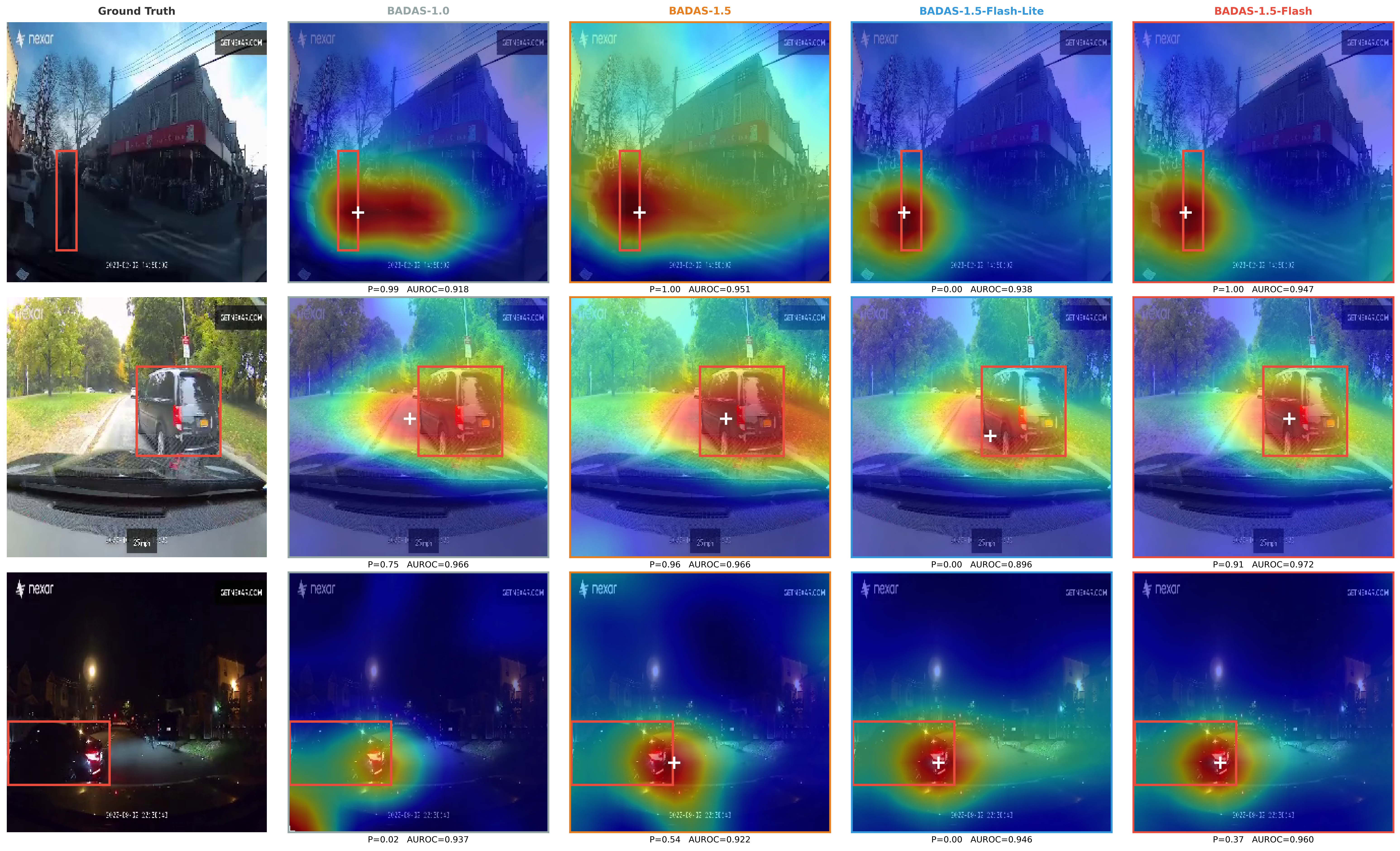}
  \caption{\textbf{Spatial attribution heatmaps.}  Each row is a driving
    scenario.  Col.~1: ground-truth frame with annotated danger region
    (orange box).  Cols.~2--5: attention heatmaps for \badasone{},
    \badas{}, \flashlite{}, and \flash{}, respectively.
    \badasone{} produces the most diffuse, scattered activations.
    The distilled models (\flashlite{}, \flash{}) show the tightest
    focus, consistently centering activation on the expected impact
    point despite having far fewer parameters.}
  \label{fig:heatmaps}
\end{figure}

Table~\ref{tab:pga} shows that the two SSL-initialized models dramatically
outperform the ViT-L models despite being 3--14$\times$ smaller.
Domain-specific SSL pre-training on Nexar driving videos builds scene
representations organized around driving-relevant regions from the outset,
while ViT-L (initialized from general video data) must simultaneously
adapt its representation and learn collision cues, yielding more diffuse attention.

\begin{table}[t]
\centering
\small
\setlength{\tabcolsep}{4pt}
\caption{\textbf{Heatmap localization (PGA).}  Random baseline: 11.5\%
  (mean GT box coverage over 1,894 clips).  SSL-initialized backbones
  achieve substantially higher localization than the ViT-L models, despite
  having 3--14$\times$ fewer parameters.}
\label{tab:pga}
\begin{tabular}{lcccc}
\toprule
\textbf{Model} & \textbf{Arch} & \textbf{SSL} & \textbf{PGA} & \textbf{$\Delta$rand} \\
\midrule
Random baseline & — & — & 11.5\% & — \\
\badasone{}     & ViT-L & $\times$ & 49.8\% & +38.3\,pp \\
\badas{}        & ViT-L & $\times$ & 52.4\% & +40.9\,pp \\
\flashlite{}    & ViT-S & $\checkmark$ & 69.8\% & +58.3\,pp \\
\flash{}        & ViT-B & $\checkmark$ & \textbf{72.1\%} & \textbf{+60.6\,pp} \\
\bottomrule
\end{tabular}
\end{table}

\subsection{BADAS-Reason: End-to-End VLM Fine-Tuning}
\label{sec:reason}

Beyond binary risk scores and heatmaps, downstream ADAS applications need
\emph{natural-language explanations}: \emph{``A pedestrian is stepping into
the lane from the right sidewalk. Brake immediately.''} We introduce
\reason{}, a four-stage pipeline for producing such explanations via a
fine-tuned VLM.

\paragraph{Stage 1 — BADAS attribution.}
We run \badas{} with attention attribution enabled on 8,680 dashcam videos
with human-authored scene descriptions.  For each video we extract the
peak-risk frame (highest predicted probability) and its associated attention
bounding boxes.

\paragraph{Stage 2 — Structured label generation.}
Each peak frame (with bounding boxes drawn) is passed to
Gemini~\cite{gemini} alongside the human-authored free-language description
written by the original annotator.  Gemini converts each human description into the structured
\texttt{\{reasoning, action\}} JSON schema — a one-sentence hazard
description and a 2--4 word driver command — yielding 6,862 valid samples
at 99.6\% parse rate.

\paragraph{Stage 3 — Dataset assembly.}
Images are prepared as \textbf{bounding-box-cropped frames} (256$\times$256, JPEG) with
bounding boxes drawn directly on the frame.  An ablation study
(Table~\ref{tab:reason}) shows that adding heatmap overlays \emph{hurts}
performance (validation loss 0.628 vs.\ 0.612), likely because the colored overlay
occludes scene details the VLM needs for semantic reasoning.  Samples are
split 80/10/10 via deterministic hash-based assignment for reproducibility.

\paragraph{Stage 4 — QLoRA fine-tuning.}
We fine-tune Qwen3-VL-4B-Instruct~\cite{qwen3vl} with QLoRA~\cite{qlora}
(rank 16, 11.8M of 4,400M parameters trainable) using supervised loss on
assistant-turn tokens, cosine LR ($1\times10^{-4}$), batch 16, 3 epochs
with early stopping, on a single NVIDIA L4 GPU (24\,GB).

\begin{table}[t]
\centering
\small
\caption{\textbf{BADAS-Reason results} on held-out test set ($n$=887).
  Ablation (bottom) compares image preparation strategies.}
\label{tab:reason}
\resizebox{\columnwidth}{!}{%
\begin{tabular}{lccc}
\toprule
\textbf{} & \textbf{Loss} & \textbf{Perplexity} & \textbf{Act.\,Match} \\
\midrule
Baseline (Qwen3-VL-4B)      & 2.657 & 14.25 & 12.2\% \\
\reason{} (fine-tuned)  & \textbf{0.612} & \textbf{1.84} & \textbf{43.6\%} \\
\midrule
\multicolumn{4}{l}{\emph{Image format ablation (val loss):}} \\
\quad Bounding-box crop (ours)       & 0.612 & — & — \\
\quad Heatmap overlay        & 0.628 & — & — \\
\bottomrule
\end{tabular}}
\end{table}

\begin{figure}[t]
  \centering
  \includegraphics[width=\linewidth]{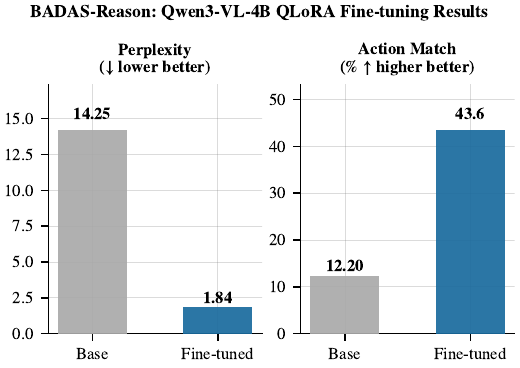}
  \caption{\textbf{\reason{} fine-tuning results.}
    QLoRA fine-tuning on the BADAS-Reason dataset reduces perplexity by
    87\% and improves action-match accuracy $3.6\times$ over the
    zero-shot Qwen3-VL-4B baseline.}
  \label{fig:reason_results}
\end{figure}

\begin{figure}[t]
  \centering
  \includegraphics[width=\linewidth]{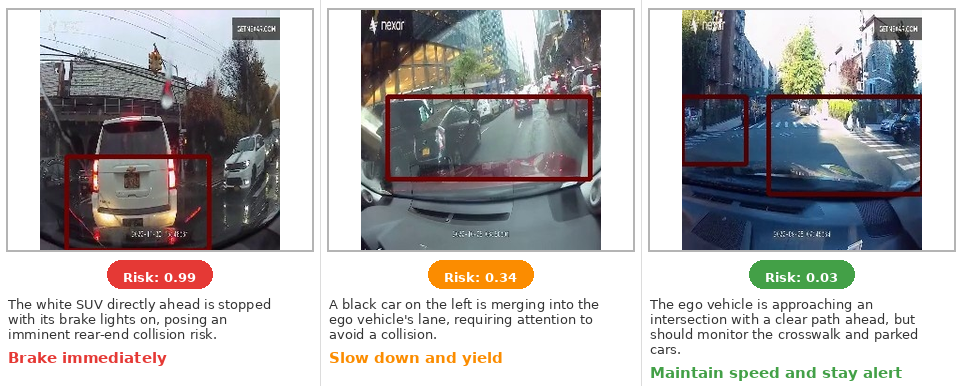}
  \caption{\textbf{BADAS-Reason output examples.}  Three representative
    clips spanning high (red, 0.99), moderate (orange, 0.34), and low
    (green, 0.03) risk.  For each clip, the pipeline overlays the BADAS
    attention bounding box on the peak-risk frame and generates a
    one-sentence hazard description and a short driver action command.
    The model produces grounded, calibrated responses across the full
    risk spectrum.}
  \label{fig:reason_examples}
\end{figure}

Figure~\ref{fig:reason_results} shows the gains.
Fine-tuning reduces perplexity by \textbf{87\%} and improves action-match
\textbf{3.6$\times$} over the zero-shot baseline.  Action-match requires
exact string match against human-authored labels and likely underestimates
semantic accuracy (annotators drew boxes around whole objects; the model
targets the impact point).  Figure~\ref{fig:reason_examples} shows three
representative outputs spanning high, moderate, and low risk, illustrating
that the model produces grounded reasoning and calibrated driver commands
across the full risk spectrum.
JSON validity is 100\% before and after fine-tuning.
The model is available to enterprise customers via the Nexar platform.

\section{Experiments}
\label{sec:experiments}

\subsection{Setup}

All models use 16 frames at 8\,fps and $256\times256$ resolution.  Sliding
window evaluation uses stride 1 (every sampled frame) for all reported
numbers.  Unless stated otherwise, classification threshold is 0.75.
GPU evaluation uses FP16 precision with ahead-of-time operator-fusion compilation.  TensorRT evaluations
on embedded platforms use FP16 engine compilation.

\subsection{Long-Tail Benchmark}
\label{sec:longtail}

Table~\ref{tab:longtail} reports per-group results on the 10-group long-tail
benchmark for all model variants.  Several group-level findings are notable.

\noindent\textbf{Infrastructure and intersection} are near-perfect for all
second-generation models (F1$\geq$0.93, EWR$\geq$0.93), indicating robust
coverage in the expanded dataset.

\noindent\textbf{Animal} remains the most challenging group: \badas{}
improves EWR from 66.1\% (\badasone{}) to 78.9\%, while simultaneously
reducing Animal FPR from 7.4\% to 3.7\% — the model detects more events
and generates fewer false alarms.  This group requires integrating subtle
motion cues over the full window, which directly penalizes smaller models:
\flashlite{} achieves 67.0\% Animal EWR vs.\ 73.4\% for \flash{}.

\noindent\textbf{Weather groups} (fog, rain, snow) show strong performance
across all models (F1$\geq$0.90), benefiting directly from the Atlas-harvested
adverse-weather clips added to the training corpus.

\noindent\textbf{False positive reduction.}
A critical practical gain of \badas{} is its dramatically lower false alarm
rate.  On the Kaggle benchmark, FPR drops from 10.9\% (\badasone{}) to
\textbf{4.6\%} (\badas{}) — a 58\% reduction — with no loss of recall.
The long-tail table shows where this matters most: infrastructure scenarios
(AUC+F1 both reach 1.000) and passing/overtaking maneuvers (100\% EWR)
were previously the two largest sources of missed detections and false
alarms.  These are precisely the scenarios enriched by the Atlas-targeted
data collection, confirming that expanding hard-negative coverage directly
translates to false positive suppression in deployment.

\begin{table*}[!t]
\centering
\small
\caption{\textbf{Long-tail benchmark results} (10 groups, 888 clips, sliding window, threshold 0.75).
  Columns: AUC, F1, Early Warning Recall (EWR), Mean Time to Alert (MTTA, in seconds).
  Best AUC, F1, and EWR per group in \textbf{bold}; ties also shown in bold.}
\label{tab:longtail}
\resizebox{\textwidth}{!}{%
\begin{tabular}{l|cccc|cccc|cccc|cccc}
\toprule
& \multicolumn{4}{c|}{\textbf{\badasone{}}}
& \multicolumn{4}{c|}{\textbf{\badas{}}}
& \multicolumn{4}{c|}{\textbf{\flash{}}}
& \multicolumn{4}{c}{\textbf{\flashlite{}}} \\
\textbf{Group} & AUC & F1 & EWR & TTA
               & AUC & F1 & EWR & TTA
               & AUC & F1 & EWR & TTA
               & AUC & F1 & EWR & TTA \\
\midrule
Animal         & 0.948 & 0.842 & 66.1\% & 0.63
               & \textbf{0.964} & \textbf{0.938} & \textbf{78.9\%} & 0.84
               & 0.948 & 0.929 & 73.4\% & 0.86
               & 0.924 & 0.881 & 67.0\% & 0.91 \\
Pedestrian     & 0.991 & 0.929 & \textbf{94.9\%} & 1.68
               & \textbf{0.998} & \textbf{0.981} & 93.7\% & 1.32
               & 0.996 & 0.957 & \textbf{94.9\%} & 1.44
               & 0.991 & 0.944 & 87.3\% & 1.57 \\
Intersection   & 0.988 & 0.803 & 94.4\% & 1.96
               & \textbf{1.000} & \textbf{0.973} & \textbf{96.3\%} & 1.76
               & \textbf{1.000} & 0.915 & \textbf{96.3\%} & 1.88
               & 0.998 & 0.931 & 92.6\% & 1.75 \\
Pass/Overtake  & 0.974 & 0.853 & 96.7\% & 1.49
               & \textbf{1.000} & \textbf{0.938} & \textbf{100\%} & 1.47
               & \textbf{1.000} & \textbf{0.938} & \textbf{100\%} & 1.67
               & 0.997 & 0.923 & 96.7\% & 1.76 \\
Cyclist        & 0.986 & 0.896 & \textbf{93.3\%} & 1.33
               & \textbf{1.000} & \textbf{0.923} & \textbf{93.3\%} & 1.16
               & 0.987 & 0.879 & \textbf{93.3\%} & 1.28
               & 0.994 & \textbf{0.923} & \textbf{93.3\%} & 1.22 \\
Motorcyclist   & 0.998 & 0.960 & 92.0\% & 1.62
               & 0.998 & \textbf{0.980} & \textbf{96.0\%} & 1.41
               & 0.998 & \textbf{0.980} & \textbf{96.0\%} & 1.51
               & \textbf{1.000} & 0.962 & \textbf{96.0\%} & 1.58 \\
Infrastructure & 0.869 & 0.807 & 68.8\% & 1.61
               & \textbf{1.000} & \textbf{1.000} & \textbf{90.6\%} & 1.47
               & 0.994 & 0.969 & \textbf{90.6\%} & 1.54
               & 0.984 & 0.939 & 87.5\% & 1.47 \\
Rain           & 0.975 & 0.868 & 94.6\% & 1.55
               & \textbf{1.000} & \textbf{0.961} & 94.6\% & 1.53
               & \textbf{1.000} & 0.937 & \textbf{97.3\%} & 1.78
               & \textbf{1.000} & 0.937 & 91.9\% & 1.76 \\
Snow           & 0.996 & 0.958 & 92.0\% & 1.40
               & \textbf{1.000} & \textbf{1.000} & \textbf{100\%} & 1.30
               & \textbf{1.000} & 0.980 & 96.0\% & 1.45
               & 0.999 & 0.980 & 96.0\% & 1.47 \\
Fog            & \textbf{1.000} & \textbf{1.000} & \textbf{100\%} & 1.44
               & \textbf{1.000} & \textbf{1.000} & \textbf{100\%} & 1.34
               & 0.998 & 0.929 & 92.9\% & 1.32
               & 0.998 & 0.929 & 92.9\% & 1.59 \\
\midrule
\textbf{Overall} & 0.949 & 0.875 & 85.5\% & 1.43
                 & \textbf{0.993} & \textbf{0.964} & \textbf{91.3\%} & 1.31
                 & 0.989 & 0.938 & 89.9\% & 1.42
                 & 0.981 & 0.927 & 85.5\% & 1.46 \\
\bottomrule
\end{tabular}}
\end{table*}

\subsection{External Benchmarks}

Table~\ref{tab:external} reports sliding-window evaluation on three
external benchmarks: DAD~\cite{dad}, DoTA~\cite{dota}, and
DADA-2000~\cite{dada2000}.
The ego-centric re-annotation protocol from \badasone{}~\cite{badas10}
is applied throughout.  We additionally include results for
Gemini-BADAS — a Gemini-2.5-Pro model fine-tuned on the BADAS
training set — as a VLM baseline.

\begin{table}[t]
\centering
\small
\caption{\textbf{External benchmark results} (sliding window, ego-centric
  evaluation).  DAD has only 13 ego-involved positives; a single missed
  detection has outsized AP impact.  Best per column in \textbf{bold}.}
\label{tab:external}
\resizebox{\columnwidth}{!}{%
\begin{tabular}{lcccccc}
\toprule
& \multicolumn{2}{c}{\textbf{DAD}} & \multicolumn{2}{c}{\textbf{DoTA}} & \multicolumn{2}{c}{\textbf{DADA-2000}} \\
\cmidrule(lr){2-3}\cmidrule(lr){4-5}\cmidrule(lr){6-7}
\textbf{Model} & AUC & AP & AUC & AP & AUC & AP \\
\midrule
\badasone{}      & 0.99 & \textbf{0.94} & 0.72 & 0.95 & 0.87 & 0.90 \\
BADAS-Open       & 0.87 & 0.66 & 0.70 & 0.94 & 0.77 & 0.87 \\
\midrule
\badas{}         & \textbf{0.993} & 0.922 & \textbf{0.991} & \textbf{0.999} & \textbf{0.991} & \textbf{0.996} \\
\flash{}         & 0.987 & 0.849 & 0.985 & 0.998 & 0.990 & 0.995 \\
\flashlite{}     & 0.982 & 0.870 & 0.985 & 0.998 & 0.981 & 0.992 \\
\midrule
Gemini-2.5-Pro   & 0.900 & 0.230 & 0.639 & 0.922 & 0.828 & 0.863 \\
Gemini-BADAS     & 0.922 & 0.659 & 0.882 & 0.976 & 0.930 & 0.953 \\
\midrule
Cosmos-BADAS     & 0.944 & 0.602 & 0.983 & 0.998 & 0.959 & 0.978 \\
Qwen3-VL-2B      & 0.754 & 0.141 & 0.709 & 0.951 & 0.805 & 0.886 \\
\bottomrule
\end{tabular}}
\end{table}

\subsection{VLM Baselines}
\label{sec:vlm_baselines}

To contextualize BADAS accuracy against modern vision-language models, we
fine-tune two state-of-the-art VLMs on the BADAS training set.

\paragraph{Cosmos-BADAS.}
NVIDIA Cosmos-Reason2-2B~\cite{cosmos} is an autoregressive video-language
model that produces free-form text rather than scalar probabilities.  We
convert it to a binary classifier via \emph{softmax over answer-token
log-probabilities}: the model is prompted with a yes/no collision question
and the probability is extracted as
\begin{equation}
  \hat{p} = \frac{\exp(\ell_A)}{\exp(\ell_A) + \exp(\ell_B)},
  \label{eq:cosmos_prob}
\end{equation}
where $\ell_A, \ell_B$ are the log-probabilities of tokens \texttt{``A''}
and \texttt{``B''} at generation position 0.

Before fine-tuning, Cosmos-Reason2-2B achieves Kaggle mAP\,0.743 but
suffers from \emph{probability compression} (collision peaks $\approx$\,0.53,
negatives $<$\,0.003), making threshold-based classification unreliable.
SFT on 479,214 clips improves substantially (Public mAP\,0.859), yet remains
\textbf{7\,pp below \badasone{}} and \textbf{14.5\,pp below \badas{}} on the
long-tail benchmark (F1\,0.817 vs.\ 0.964), suggesting V-JEPA2's dense
temporal prediction better suits physical collision anticipation.

\paragraph{Gemini-BADAS.}
Gemini~\cite{gemini} exposes no gradient access; we fine-tune via Vertex AI
SFT on 17,479 BADAS-Reason video--description pairs across multiple safety
tasks to mitigate catastrophic forgetting.  Since log-probabilities are
unavailable, the model outputs a decimal in $[0,1]$; we average over three
temperatures to reduce variance:
\begin{equation}
  \hat{p} = \tfrac{1}{3}(p_{0.0} + p_{0.3} + p_{0.7}).
  \label{eq:gemini_ensemble}
\end{equation}
Single-window accuracy improves from F1\,0.571 (vanilla) to F1\,0.662
(tuned); sliding-window results appear in Table~\ref{tab:external}.

\subsection{Inference Latency}

\begin{figure}[t]
  \centering
  \includegraphics[width=\linewidth]{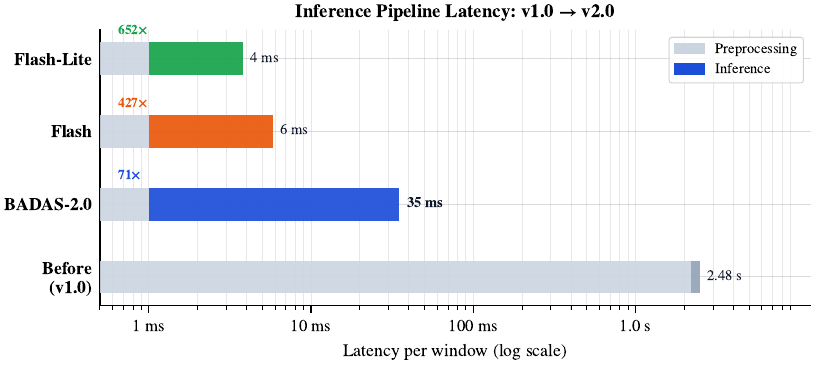}
  \caption{\textbf{Inference pipeline latency: v1.0 \textrightarrow\ v2.0.}
    End-to-end latency decomposed into preprocessing (light) and inference
    (dark).  Total latency drops from 2.5\,s to 35\,ms ($71\times$).}
  \label{fig:latency_opt}
\end{figure}

\paragraph{Optimization journey from v1.0 to v2.0.}
Figure~\ref{fig:latency_opt} compares end-to-end latency for each model variant,
decomposed into preprocessing (light) and inference (dark).
\badasone{} ran at \textbf{2.5\,s} per window (preprocessing 2200\,ms, FP32
inference 277.6\,ms).  Four targeted optimisations bring \badas{} to
\textbf{35\,ms total} ($\mathbf{71\times}$ faster): FP16 quantisation and
ahead-of-time graph compilation reduce inference from 277.6\,ms to 33.4\,ms;
a streaming rolling buffer and GPU-native preprocessing shrink the preprocessing
stage from 2200\,ms to under 1\,ms.

The same GPU pipeline is shared across the entire model family.
Table~\ref{tab:latency} extends the comparison to \flash{} and \flashlite{}
across both platforms; \flash{} and \flashlite{} deliver a further
$7$--$12\times$ reduction through architecture distillation, all within
the 125\,ms real-time budget.

\begin{table}[h!]
\centering
\small
\setlength{\tabcolsep}{2.5pt}
\caption{\textbf{Inference latency per prediction window.}  FP16 precision
  with ahead-of-time compilation (end-to-end, $+$0.4--0.7\,ms preprocessing).
  \badasone{} measured on A100 FP32 with full-video decode.}
\label{tab:latency}
\resizebox{\columnwidth}{!}{%
\begin{tabular}{lcrrr}
\toprule
\textbf{Model} & \textbf{Params}
  & \textbf{A100} & \textbf{J.\,Thor} & \textbf{Speedup} \\
\midrule
\badasone{}    & 300M & 2500\,ms & --- & --- \\
\midrule
\badas{}       & 300M & 34\,ms & 41\,ms & 74$\times$ \\
\flash{}       & 86M  & 4.8\,ms & 12.5\,ms & 521$\times$ \\
\flashlite{}   & 22M  & 2.8\,ms & 5.9\,ms  & 893$\times$ \\
\bottomrule
\end{tabular}}
\end{table}

All variants fit the 125\,ms budget on Jetson Thor; preprocessing adds only
0.4--0.7\,ms per window, pipelined asynchronously with inference.

\begin{figure}[t]
  \centering
  \includegraphics[width=\linewidth]{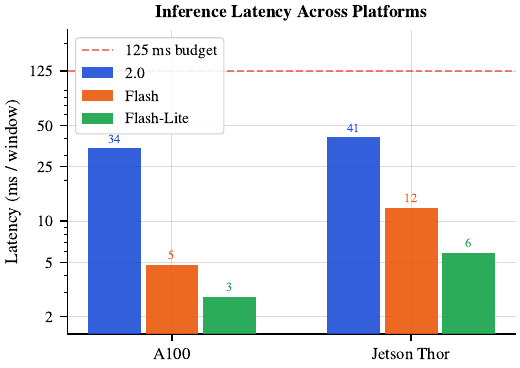}
  \caption{\textbf{Inference latency per prediction window} across three
    deployment platforms (log scale).  Dashed line: 125\,ms real-time
    budget at 8\,fps.  All \badas{} variants satisfy the budget on
    both platforms; \flashlite{} at 2.8\,ms on A100 leaves 44$\times$ headroom.}
  \label{fig:latency}
\end{figure}

\section{Conclusion}

BADAS-2.0 demonstrates that collision anticipation is, at its core, a data
and representation problem rather than an architectural one.  The gains
reported here did not come from a novel loss function or a larger backbone
— they came from closing three systematic gaps that BADAS-1.0 left open:
a training distribution blind to rare events, an inference cost incompatible
with edge hardware, and a risk score that told operators \emph{when} but
not \emph{why}.

The most transferable lesson is the compounding value of a deployed model.
Once BADAS-1.0 was in production, it became the cheapest annotator in the
pipeline: its risk scores surfaced hard cases that human labelers would never
encounter through random sampling, its attention maps provided ready-made
supervision for \reason{}, and its calibrated uncertainty transferred directly
to the distilled students through soft targets.  That last point has a
prerequisite, however: the students could only absorb that uncertainty because
domain-specific SSL had already aligned their representations to the dashcam
distribution.  The gap between a randomly initialised ViT-S (AP~0.693) and
its SSL-pretrained counterpart (AP~0.974) is larger than the gap between any
two supervised models in this paper — a result with direct implications for
how edge variants of future systems should be developed.  Future systems that
neglect either half of this loop — the deployed oracle or the domain-adapted
student — leave substantial efficiency on the table.

Inference code and evaluation benchmarks are publicly available at the
project page.

\paragraph{Limitations and future work.}
Animal EWR remains below 80\% even for the largest model.  A share of this
gap is irreducible: animals frequently enter the roadway with near-zero
warning time regardless of detection quality, and the fundamental constraint
is scenario geometry rather than model capacity.  The remaining gap points
toward richer temporal context and multi-frame feature aggregation as the
most promising directions.

Beyond accuracy, three architectural directions stand out.

\textbf{Semantic spatial grounding.}  The current attention heatmaps
operate at patch resolution and are extracted as a post-hoc proxy.
Replacing them with a pixel-aligned segmentation head — trained jointly
with the collision objective — would promote each predicted region from
a saliency highlight to a typed object instance.  Downstream components
could then condition on object class and extent rather than on a continuous
activation map, enabling more precise handoff to planning and intervention
modules.

\textbf{Trajectory-conditioned risk.}  BADAS-2.0 outputs a scalar
collision probability but does not model \emph{which evasive action removes
the risk}.  Augmenting the prediction head to jointly estimate a
safety-restoring trajectory — braking profile, steering correction, or
lane change — would close the loop between anticipation and action,
making the system directly actionable rather than purely alerting.

\textbf{Intrinsic reasoning from model representations.}
\reason{} currently externalises explanation to a separate VLM that consumes
attention heatmaps as a proxy for model state.  A tighter design would
extract textual reasoning directly from the encoder's latent
representations — bypassing the heatmap intermediary and the modality
gap it introduces — yielding explanations that are causally grounded in
the same features that drove the risk score.

{\small
\bibliographystyle{ieee_fullname}
\bibliography{refs}
}
\end{document}